%% file: main.tex
\documentclass[conference]{IEEEtran}
\usepackage{cite}
\usepackage{amsmath,amssymb,amsfonts}
\usepackage{algorithmic}
\usepackage{graphicx}
\usepackage{textcomp}
\usepackage{xcolor}
\usepackage{subfig}
\usepackage{multicol}
\usepackage{multirow}
\usepackage{booktabs}
\usepackage{bm}
\usepackage{hyperref}
\usepackage{enumitem}
\usepackage[labelfont=bf]{caption}
\usepackage{tikz}
\usepackage{textcomp}
\usepackage{lipsum}
\usepackage{dblfloatfix}
\usepackage{comment}

\DeclareMathOperator{\sign}{sign}

\DeclareMathOperator{\MSE}{MSE}

\def\BibTeX{{\rm B\kern-.05em{\sc i\kern-.025em b}\kern-.08em
    T\kern-.1667em\lower.7ex\hbox{E}\kern-.125emX}}
    
\makeatletter
\newcommand*\titleheader[1]{\gdef\@titleheader{#1}}
\AtBeginDocument{%
  \let\st@red@title\@title
  \def\@title{%
    \bgroup\normalfont\normalsize\centering \vspace{-25pt}\@titleheader\par
    \egroup
    \vspace{13pt}
    \st@red@title
    }
    
}
\makeatother

\title{Adversarial Perturbations Fool Deepfake Detectors}
\titleheader{To appear in the proceedings of the \textit{International Joint Conference on Neural Networks} (IJCNN 2020).}
\author{\IEEEauthorblockN{Apurva Gandhi}
\IEEEauthorblockA{
\textit{Department of Electrical and Computer Engineering} \\
\textit{University of Southern California}\\
Los Angeles, USA \\
apurvaga@usc.edu}
\and
\IEEEauthorblockN{Shomik Jain}
\IEEEauthorblockA{
\textit{Department of Electrical and Computer Engineering} \\
\textit{University of Southern California}\\
Los Angeles, USA \\
shomikja@usc.edu}}

\newcommand\copyrighttext{%
  \footnotesize \textcopyright 2020 IEEE. Personal use of this material is permitted. Permission from IEEE must be obtained for all other uses, in any current or future media, including reprinting/republishing this material for advertising or promotional purposes, creating new collective works, for resale or redistribution to servers or lists, or reuse of any copyrighted component of this work in other works.}
\newcommand\copyrightnotice{%
\begin{tikzpicture}[remember picture,overlay]
\node[anchor=south,yshift=20pt] at (current page.south) {\fbox{\parbox{\dimexpr\textwidth-\fboxsep-\fboxrule\relax}{\copyrighttext}}};
\end{tikzpicture}%
}

\begin{document}

\maketitle
\copyrightnotice

\begin{abstract}
This work uses adversarial perturbations to enhance deepfake images and fool common deepfake detectors. We created adversarial perturbations using the Fast Gradient Sign Method and the Carlini and Wagner L$\bm{_2}$ norm attack in both blackbox and whitebox settings. Detectors achieved over 95\% accuracy on unperturbed deepfakes, but less than 27\% accuracy on  perturbed deepfakes. We also explore two improvements to deepfake detectors: (i) Lipschitz regularization, and (ii) Deep Image Prior (DIP). Lipschitz regularization constrains the gradient of the detector with respect to the input in order to increase robustness to input perturbations. The DIP defense removes perturbations using generative convolutional neural networks in an unsupervised manner. Regularization improved the detection of perturbed deepfakes on average, including a 10\% accuracy boost in the blackbox case. The DIP defense achieved 95\% accuracy on perturbed deepfakes that fooled the original detector while retaining 98\% accuracy in other cases on a 100 image subsample.
\end{abstract}

%Regularization improved the detection of perturbed deepfakes in all cases including a 24\% accuracy boost on average in the blackbox case.


\section{Introduction}
\begin{figure}[t!] 
    \centering
    \includegraphics[width=\linewidth]{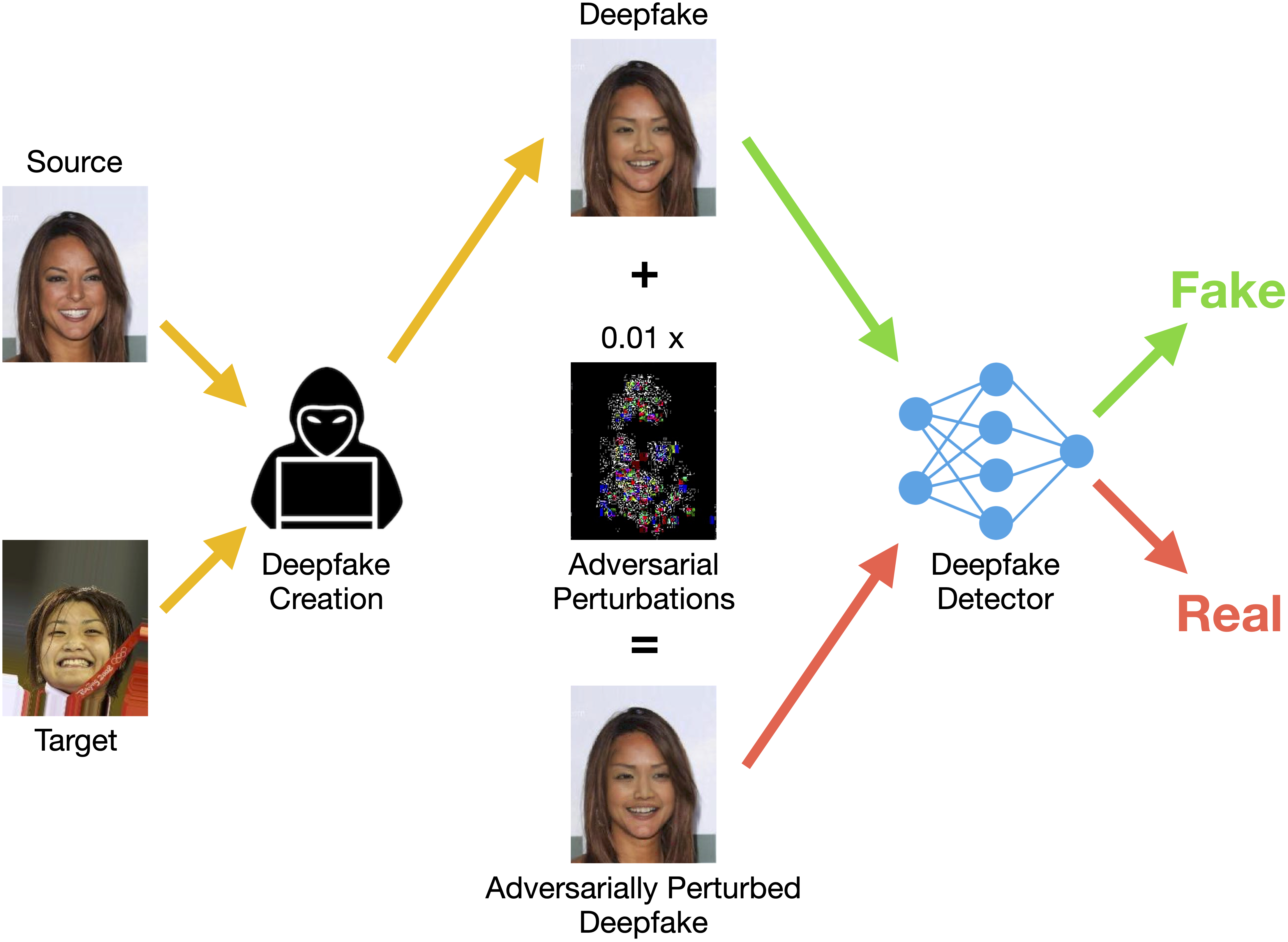}
    \caption{\textbf{Deepfake creation and adversarial perturbation.} Example of creating a deepfake from source and target faces and adding adversarial perturbations to it. A deepfake detector correctly classifies the deepfake as fake but classifies the adversarially perturbed deepfake as real.  
    }
    \label{figure1}
\end{figure}

% (derived from ``deep learning” and ``fake”)
This work enhances deepfakes with adversarial perturbations to fool common deepfake detectors. \textit{Deepfakes} replace a ``source" individual in an image or video with a ``target" individual's likeness using deep learning \cite{df_review}. \textit{Adversarial perturbations} are modifications made to an image in order to fool a classifier. An adversary can choose these perturbations to be small so that the difference between the perturbed and original images is visually imperceptible. Figure~\ref{figure1} shows a deepfake generated from source and target images as well as its adversarially perturbed version. A deepfake detector correctly classifies the original as fake but fails to detect the perturbed deepfake which looks almost identical. In our results, detectors achieved over 95\% accuracy on unperturbed deepfakes, but less than 27\% accuracy on perturbed deepfakes. 

%even with limited training data on the target individual 
% Previous research has shown that adversarial attacks successfully reduce the accuracy of models in a variety of domains, including for the MNIST, CIFAR-10, and Image-Net datasets \cite{fgsm}. These malicious uses of deepfakes violate individuals’ identity and privacy and can also propagate misinformation in society. Deepfakes’ harms make detection methods imperative, especially on social media where deepfakes disseminate quickly

Deepfakes have been used for many malicious applications. In 2019, an app called DeepNude was released which could take an image of a fully-clothed woman and generate an image with her clothes removed \cite{deepnude}. Furthermore, Facebook found over 500 accounts spreading pro-President Trump, anti-Chinese government messages using deepfake profile pictures  \cite{cnn_profle}. These harmful uses of deepfakes violate individuals’ identity and can also propagate misinformation, especially on social media. Ahead of the 2020 U.S. election, Facebook and Twitter stated plans to try and remove certain deepfakes \cite{twitter_facebook}. But adversarial perturbations can compromise the performance of deepfake detection methods used on these platforms.  

%BuzzFeed also recently released a now-infamous fake video of President Obama saying profanity-ridden and derogatory statements \cite{obama_video}. The advancement of deep learning models such as generative adversarial networks (GANs) and autoencoders has made it easier to create deepfakes \cite{df_review}.

To defend against these perturbations, we explore two improvements to deepfake detectors: (i) Lipschitz regularization, and (ii) Deep Image Prior. \textit{Lipschitz regularization}, introduced in \cite{regularization}, constrains the gradient of the detector with respect to the input data. We use \textit{Deep Image Prior} (DIP), originally an image restoration technique \cite{dip}, to remove perturbations by iteratively optimizing a generative convolutional neural network in an unsupervised manner. To our knowledge, this is the first application of DIP for removing adversarial perturbations. Overall, the contributions of this work aim to highlight the vulnerability of deepfake detectors to adversarial attacks, as well as present methods to improve robustness.

\section{Deepfake Creation and Detection}

\begin{figure*}[t!]
\centering
\mbox{\subfloat[Unperturbed Real Images]{
\includegraphics[width=0.42\linewidth]{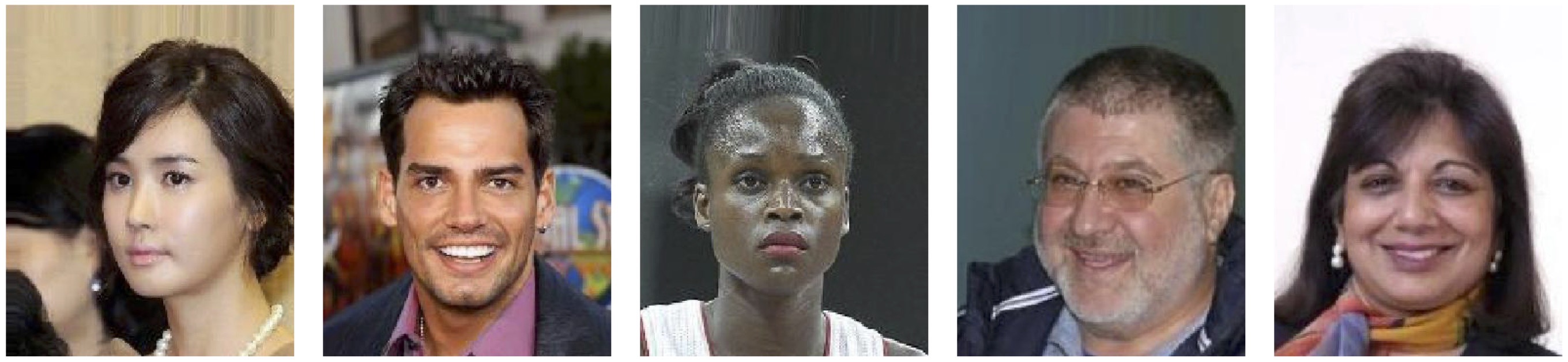}}\qquad\quad

\subfloat[Unperturbed Fake Images]{
\includegraphics[width=0.42\linewidth]{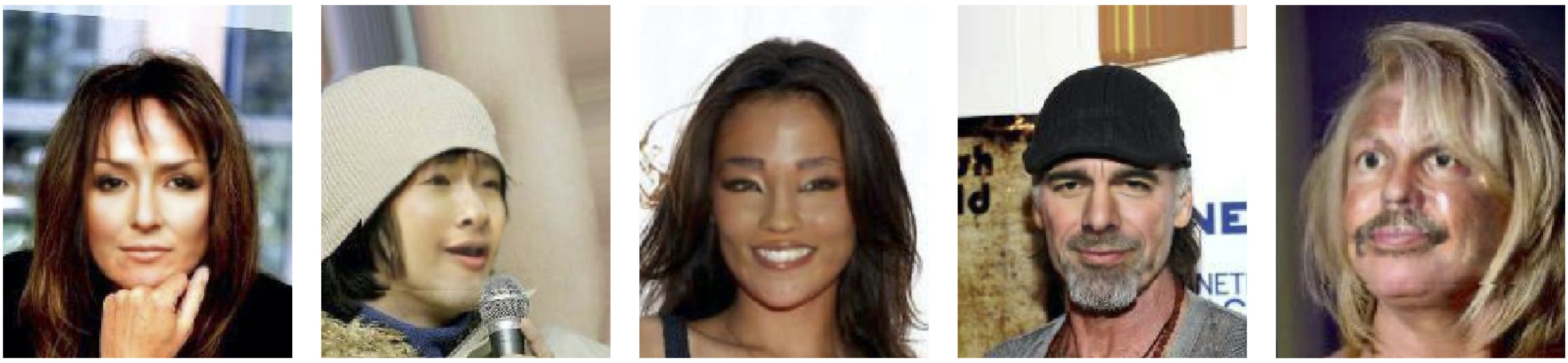}}
}

\mbox{
\subfloat[Perturbed (FGSM) Fake Images]{
\includegraphics[width=0.42\linewidth]{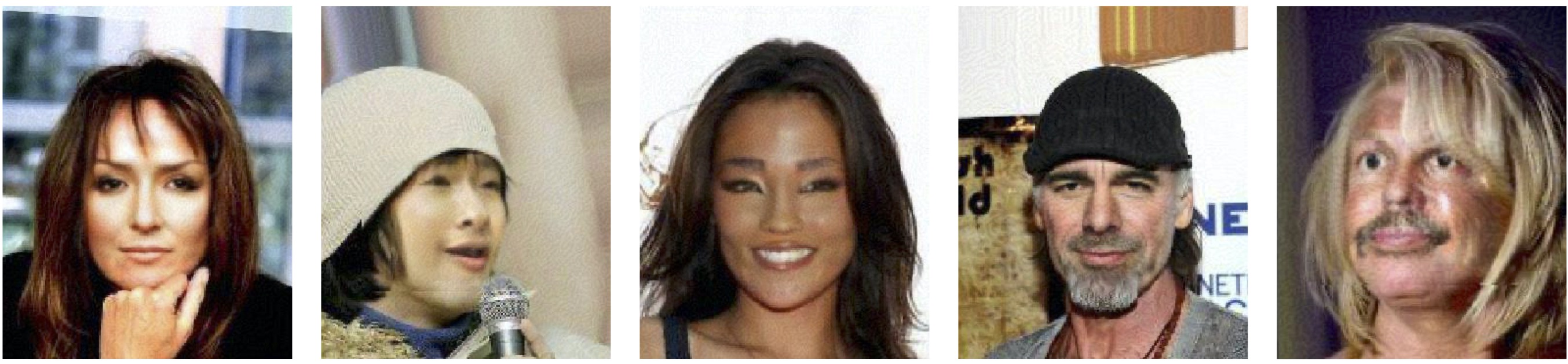}}\qquad\quad

\subfloat[Perturbed (CW-$\textit{L}_{2}$) Fake Images]{
\includegraphics[width=0.42\linewidth]{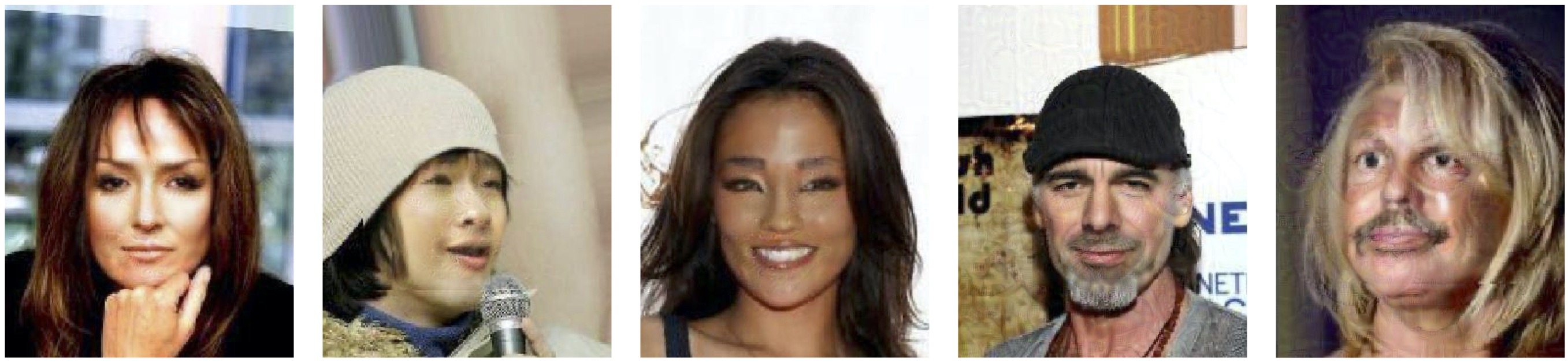}}}

\caption{\textbf{Examples of real, fake, and perturbed images.} Adversarially perturbed deepfakes (c and d) look similar to unperturbed deepfakes (b). Some fake images look more realistic than others. We use the ResNet model to perturb the first 3 fake images and the VGG model to perturb the last 2 fake images. All adversarial examples shown fool both models.}
\label{dataset_sample}
\end{figure*}

%Indeed, concurrent work has tried extending adversarial peturbations to deepfake videos \cite{neekhara2020adversarial}. 
We focus on deepfake images of celebrity faces as the scope of this work. Our dataset consists of 10,000 images: 5,000 real and 5,000 fake. The 5,000 real images were randomly sampled from the CelebA dataset \cite{celeba}. Fig.~\ref{dataset_sample} includes examples of real and fake images from our dataset. 

\subsection{Deepfake Creation}
Most deepfake creation methods use generative adversarial networks (GAN) to replace the face of a ``source" individual with that of a ``target" individual \cite{df_review}. The generator in these methods consists of an encoder-decoder based network. First, the methods train a common encoder but different decoders for each face. Then, the source image is passed through the common encoder and the target’s decoder to create a deepfake. A shortcoming of these methods is that training the encoder and decoder networks requires many images of both the source and target individuals. Creating a dataset of deepfakes using these methods is difficult: We would require numerous images for each individual and would also have to train separate decoders for each target. 

%Most deepfake creation methods use a generative adversarial network (GAN) \cite{goodfellow_gan, schmidhuber_gan} to replace the face of a ``source" individual with that of a ``target" individual \cite{df_review}. Within the generator, these methods use a common encoder but separate decoder for 
Instead, we created the 5,000 fake images in our dataset using an existing implementation called the ``Few-Shot Face Translation GAN" \cite{fewshot_ftg}. This implementation takes inspiration from Few-Shot Unsupervised Image-to-Image Translation (FUNIT) \cite{funit} and Spatially-Adaptive Denormalization (SPADE) \cite{spade}. FUNIT transforms an image from a source domain to look like an image from a target domain. Moreover, it does so using only a single source image and a small set of (or even a single) target image. It achieves this by simultaneously learning to translate between images sampled from numerous source and target domains during training; this allows FUNIT to generalize to unseen source and target domains at test time \cite{funit}. SPADE is a normalization layer that conditions normalization on an input image segmentation map in order to preserve semantic information \cite{spade}. The Few-Shot Face Translation GAN adds SPADE units to the FUNIT generator allowing us to create a deepfake using only a single source and target image.

\subsection{Detection Methods}\label{sec_detection}
%CNNs are effective for detecting the resolution inconsistency between the warped target face and surrounding context from the source image characteristic of many deepfake generation methods \cite{df_warped}. 

Common deepfake detection methods use convolutional neural networks (CNNs) to classify images as ``real" or ``fake" \cite{df_review}. Prior work has shown that the VGG \cite{vgg} and ResNet \cite{resnet} CNN architectures achieve high accuracy for detecting deepfakes from a variety of creation methods \cite{df_review}. The original architectures for both these models have thousand-dimensional output vectors. Instead, we replaced the last layer of these architectures to output two-dimensional softmax vectors corresponding to the real and fake classes. We chose a softmax vector over a sigmoid scalar to make the models compatible with the Carlini and Wagner $L_2$ norm attack discussed in section~\ref{sec_cw}.

%To make them suitable for binary classification, we replaced the last layer of the networks to output two-dimensional vectors instead of the thousand-dimensional vectors in the original architectures. While a one-dimensional output would suffice for binary classification, we used a two-dimensional softmax output to make the models compatible with the Carlini and Wagner $L_2$ norm attack discussed in section III-B. 

We tested the VGG-16 and ResNet-18 architectures on our dataset. The models achieved train accuracies of 99.9\% and 94.7\% as well as test accuracies of 99.7\% and 93.2\%, respectively. These results are based on a 75\%-25\% train-test split after 5 epochs of training with a batch size of 16. Table~\ref{detection_results} reports Area Under the Receiver Operating Characteristic (AUROC) curve values and additional performance metrics for these deepfake detectors.

% Moreover, concurrent work has shown that perturbing less than 1\% of a deepfake's pixels can render a detector unusable \cite{carlini2020evading}. 
\section{Adversarial Attacks}
Deep neural networks and many other pattern recognition models are vulnerable to adversarial examples -- input data that has been perturbed to make the model misclassify the input \cite{carlini2017adversarial}. The adversary can further craft these \textit{adversarial perturbations} to have small magnitude so that the adversarial examples are difficult to distinguish from the original unperturbed input data. We tested the effect of adversarial perturbations on deepfake detectors using the following two attacks: the Fast Gradient Sign Method (FGSM) \cite{fgsm} and the Carlini and Wagner $\textit{L}_{2}$ Norm attack (CW-$\textit{L}_{2}$) \cite{cw}. We chose FGSM to try a popular, efficient attack and CW-$\textit{L}_{2}$ to try a slow but stronger attack. This work only considers perturbations of fake images: An adversary's goal is to manipulate deepfakes so that they are classified as real and not vice versa. Concurrent work has extended adversarial perturbations of deepfakes with additional attacks \cite{carlini2020evading} and even to videos \cite{neekhara2020adversarial}.

\subsection{Fast Gradient Sign Method (FGSM)}
Let $\textbf{x}$ be the vector of pixel values in an input image and $\textbf{y}$ be the corresponding true target class value. Let $J(\textbf{x}, \textbf{y}, \theta)$ be the training loss function (e.g. categorical cross-entropy loss for a softmax classifier) where $\theta$ represents the parameters of the model. FGSM exploits the gradient of the loss with respect to the input, $\nabla_xJ(\textbf{x}, \textbf{y}, \theta)$, to generate the adversarial example, $\textbf{x}_{adv}$:
\begin{equation}
    \textbf{x}_{adv} = \textbf{x} + \epsilon\,\sign(\nabla_xJ(\textbf{x}, \textbf{y}, \theta)).
\end{equation}

Here, $\epsilon$ is a hyperparameter that controls the magnitude of the perturbation per pixel. By keeping $\epsilon$ small, we can limit the magnitude of the perturbations and thus minimize visual distortions in the adversarial examples. In practice, the pixel values of the adversarial examples are further clipped to a range of floating point values between 0 and 1. We used an $\epsilon$ value of 0.02 to generate our FGSM adversarial examples. This value was chosen after evaluating the attack effectiveness and visual distortions for several $\epsilon$ values in the range [0.01, 0.10]. 

To see why this attack is effective in causing a misclassification, we examine the linear approximation of the loss using its Taylor series expansion:
\begin{multline}
      J(\textbf{x}_{adv}, \textbf{y}, \theta) \approx J(\textbf{x}, \textbf{y}, \theta) \\+ \epsilon\,\nabla_xJ(\textbf{x}, \textbf{y}, \theta)^T\sign(\nabla_xJ(\textbf{x}, \textbf{y}, \theta)).
\label{eq_taylor}
\end{multline}

Using the $\sign$ function of the gradient ensures that the dot product in the second term of (\ref{eq_taylor}) is non-negative. Thus, FGSM chooses the perturbation that causes the maximum increase in the value of the linearized loss function subject to the $\epsilon$ pixel-perturbation control parameter.

\subsection{Carlini and Wagner $\textit{L}_{2}$ Norm Attack (CW-$\textit{L}_{2}$)}\label{sec_cw}
This attack simultaneously minimizes two objectives. Let \textbf{x'} be a perturbed image. The first objective is to minimize the $\textit{L}_{2}$ norm of the perturbation:
\begin{equation}
    \underset{x'}{\min}\{\|\textbf{x'}- \textbf{x}\|_{2}^2\}.
\label{eq_cw1}
\end{equation}
The second objective tries to make the perturbation cause a misclassification. Let $\textbf{Z}(\textbf{x})$ represent the pre-softmax vector output (or logits) of a multi-class neural network classifier. The second objective is as follows:
\begin{equation}
    \begin{gathered}
    \underset{x'}{\min}\{f(\textbf{x'})\} \\
    \text{where} \ f(\textbf{x'}) = \max(\underset{i \neq y}{\max}\{\textbf{Z}(\textbf{x'})_y - \textbf{Z}(\textbf{x'})_i\}, -\kappa). 
    \end{gathered}
\label{eq_cw2}
\end{equation}
Here, $i$ and $y$ index into $\textbf{Z}(\textbf{x'})$ with $y$ being the index of the true target class. By minimizing $f(\textbf{x'})$, we try to maximize the difference between the logit of an incorrect class and the logit of the true class. Since the predicted class corresponds to the maximum logit, minimizing $f(\textbf{x'})$  effectively tries to cause a misclassification. $\kappa$ is a parameter that defines a threshold by which the logit corresponding to the incorrect predicted class should exceed the logit of the true target class. 

%Since the class corresponding to the maximum logit is chosen as the predicted class,

The attack also performs a change of variable from \textbf{x'} to $\bm{\omega}$:
\begin{equation}
    \textbf{x'} = \frac{1}{2}(\tanh(\bm{\omega}) + 1).
\label{eq_cw3}
\end{equation}
This ensures that the perturbed image ($\textbf{x'}$) has floating point pixel values between 0 and 1. Putting (\ref{eq_cw1}), (\ref{eq_cw2}) and (\ref{eq_cw3}) together, we obtain the CW-$\textit{L}_{2}$ attack:
\begin{equation}
    \begin{gathered}
     \bm{\omega}^* = \underset{\omega}{\arg\min}\{\|\textbf{x'} - \textbf{x}\|_{2}^2 + c \, f(\textbf{x'}) \} \\ 
     \textbf{x}_{adv} = \frac{1}{2}(\tanh(\bm{\omega}^*) + 1).
    \end{gathered}
\end{equation}
$c$ is positive and controls the relative strength of the two objectives. In practice, $c$ is chosen using a modified binary search which finds the smallest value of $c$ in a provided range, such that $f(\textbf{x}_{adv})$ is less than 0. This search along with the iterative gradient descent optimization process makes the attack very slow. However, this attack breaks many previously proposed defenses against adversarial examples \cite{carlini2017adversarial}. For further details about the attack, we refer the reader to the CW-$\textit{L}_{2}$ paper \cite{cw} and the implementation we used \cite{cw_implementation}. 

For all adversarial examples generated using this method, we chose $[10^2, 10^4]$ as the range for $c$ with 5 search steps. We performed a maximum of 1000 iterations for optimization with a learning rate of $0.01$. We used 200 for the value of $\kappa$. The value of $\kappa$ was chosen by trying out values in the range $[0, 500]$. The range of $c$ was chosen by initially performing attacks using a range of $[10^{-10}, 10^{10}]$ and then narrowing down the range to include the values of $c$ most commonly chosen by the search steps. The values for $\kappa$ and range of $c$ were evaluated objectively based on the decrease in accuracy of the classifier under attack and subjectively based on the amount of visible distortions in the perturbed images. We left all other parameters to the defaults recommended by the implementation \cite{cw_implementation}. 

\input{tables/detection_results.tex}

\input{tables/attack_results.tex}

\subsection{Attack Types}
Adversarial attacks on machine learning models fall into two types depending on the amount of information available to the adversary about the model under attack: 
\begin{itemize}[leftmargin=*]
    \item \textit{Whitebox Attack}:
    The adversary has complete access to the model under attack, including the model architecture and parameters. It may be unlikely for an adversary to have access to model parameters in many scenarios. However, machine learning solutions such as deepfake detectors often use existing, publicly known and accessible architectures for transfer learning purposes \cite{df_review}. 
    \item \textit{Blackbox Attack}: The adversary has limited or almost no information about the model under attack. Previous research \cite{szegedy2013intriguing, fgsm, papernot2016transferability} has shown that adversarial examples created using whitebox attacks on one model also damage performance of different models trained for the same task. Furthermore, these attacks do not even have to be in the same family of classifiers. For example, adversarial examples created using a neural network also work on support vector machines and decision tree classifiers \cite{papernot2016transferability}. This \textit{transferability} of adversarial examples is what makes blackbox attacks possible. Blackbox attacks can involve varying degrees of access to the model under attack, such as access to the predicted probabilities, predicted class or even the training data \cite{carlini2019evaluating}. This work assumes the last of these and performs blackbox attacks on the VGG model by creating whitebox examples for the ResNet model. Similarly, blackbox examples for the ResNet model are generated by creating whitebox examples for the VGG model. We note that since our models obtained over 94\% training accuracy, access to ground-truth class information is almost equivalent to access to only the predicted class information. 
\end{itemize}

% Blackbox attacks can be conducted assuming various degrees of access to the model under attack (e.g. access to predicted probabilities, predicted class, or even the training data) \cite{carlini2019evaluating}. In this work, we assume the last of these, and perform blackbox attacks on the VGG model by creating whitebox examples for the ResNet model. Similarly, blackbox examples for the ResNet model are generated by creating whitebox examples for the VGG model. We note that since our models achieve greater than 99.9\%  training accuracy, access to ground-truth class information is virtually equivalent to access to only predicted class information.  

\begin{figure}[b!] 
    \centering
    \includegraphics[width=\linewidth]{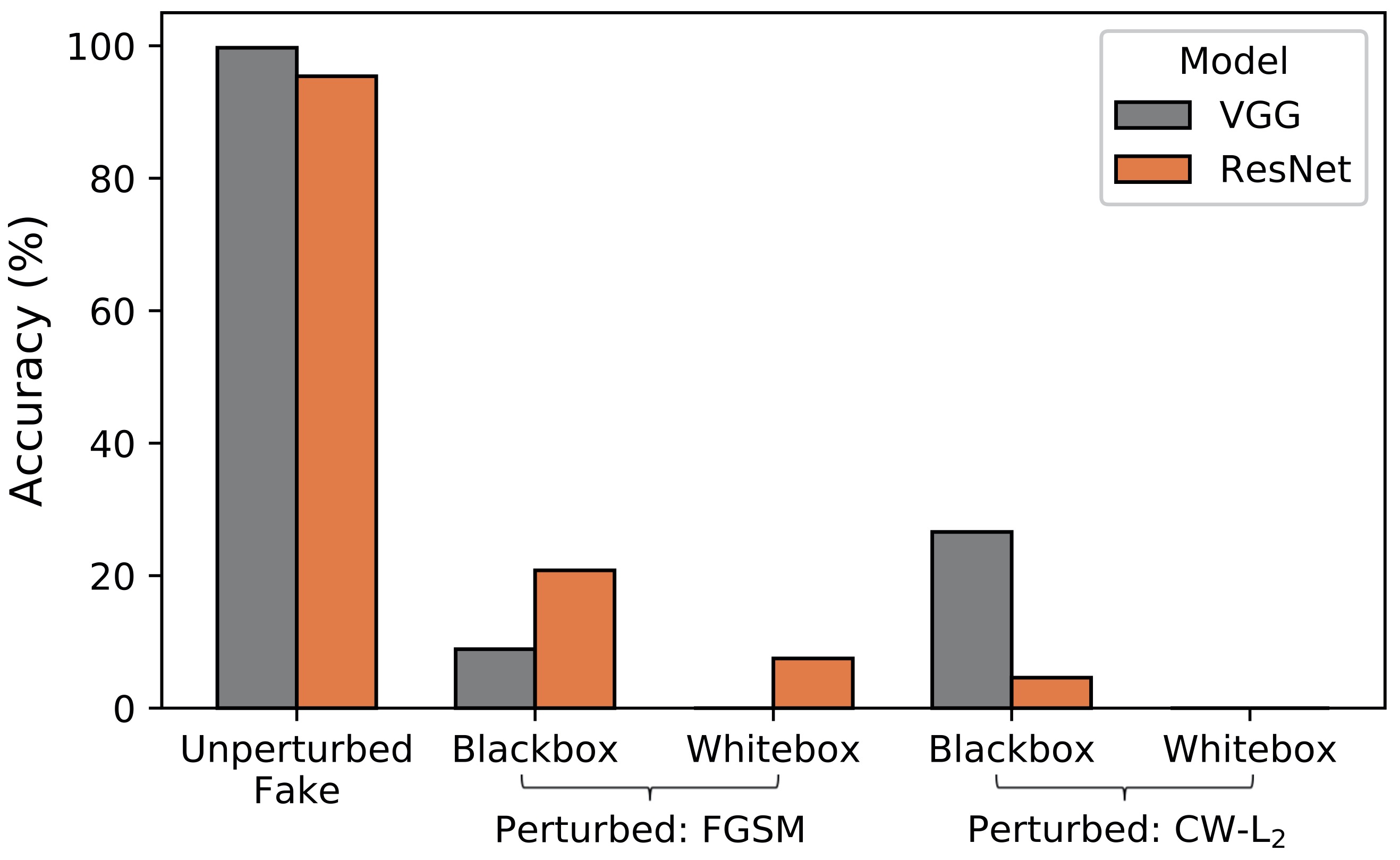}
    \caption{\textbf{Adversarial attack results.} VGG and ResNet accurately detected unperturbed fake images but performed significantly worse on adversarially perturbed fake images. Blackbox attacks were less effective than whitebox attacks. CW-$\textit{L}_{2}$ was generally more effective than FGSM.}
    \label{attack_results}
\end{figure}

\subsection{Attack Results}\label{sec_attack_results}

Adversarial attacks significantly reduced the performance of both the VGG and ResNet deepfake detection models. We compare results on datasets of unperturbed and perturbed fake images created using the test set. The datasets exclude real images since they were not perturbed. Fig.~\ref{attack_results} and Table~\ref{attack_results_table} show the adversarial attack results.

\begin{figure}[b!] 
    \centering
    \includegraphics[width=0.9775\linewidth]{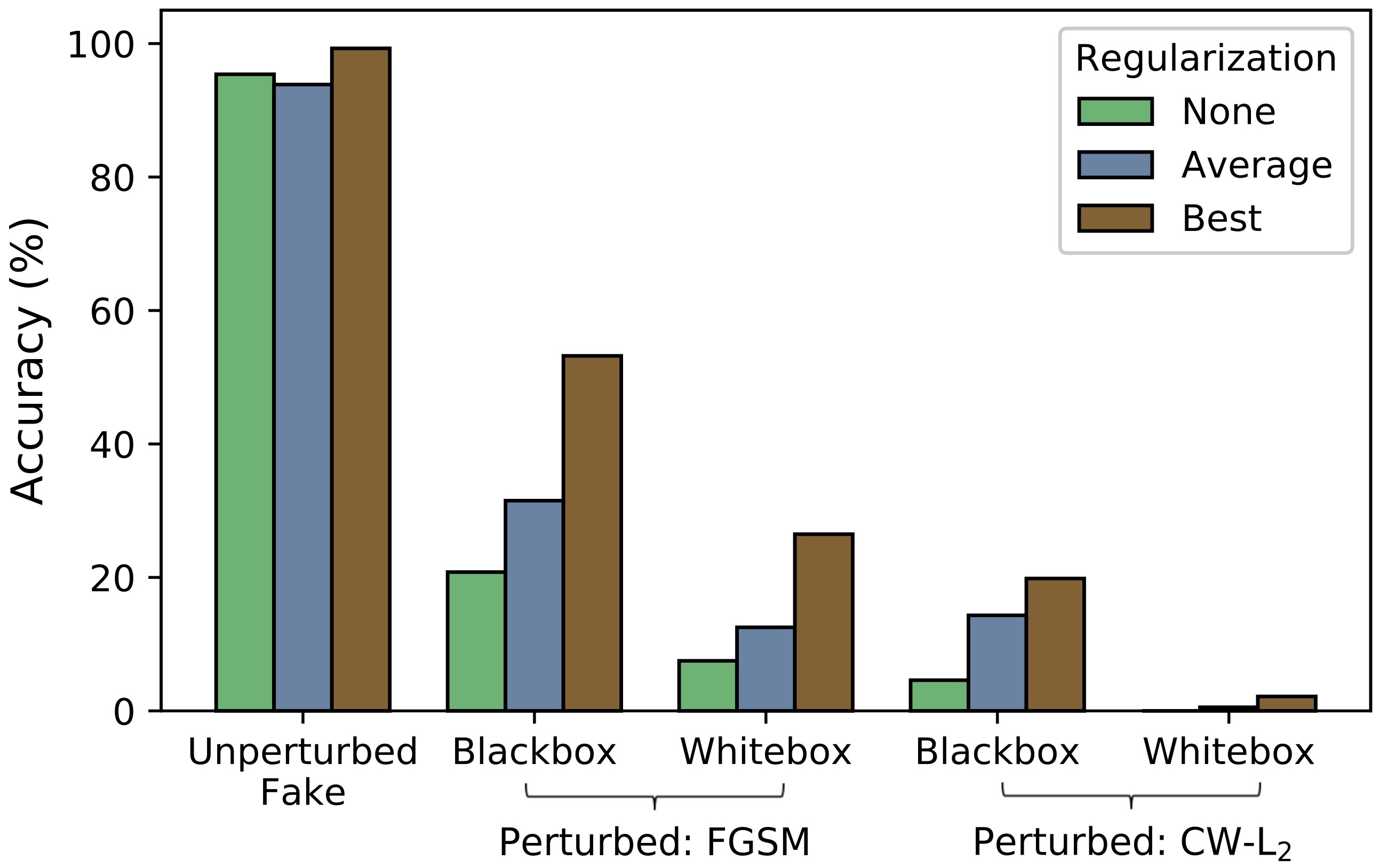}
    \caption{\textbf{Regularization results.} Regularization improved the detection of adversarially perturbed deepfakes overall. Regularization mostly maintained accuracy on unperturbed images. Results are plotted for the average and best performances among models with varying regularization strength.}
    \label{regularization_results}
\end{figure}

For unperturbed fake images, VGG achieved an accuracy of 99.7\% and ResNet achieved 95.2\%. In the blackbox FGSM case, the accuracy decreased to 8.9\% for VGG and 20.8\% for ResNet. Blackbox CW-$\textit{L}_{2}$ reduced the accuracy of VGG to 26.6\% and ResNet to 4.6\%. In the whitebox FGSM case, the accuracy dropped to 0.0\% for VGG and 7.5\% for ResNet. Whitebox CW-$\textit{L}_{2}$ lowered the accuracy of both VGG and ResNet to 0.0\%. As in section~\ref{sec_detection}, these results are based on a 75\%-25\% train-test split.

Whitebox attacks were more effective than blackbox attacks. This is expected because whitebox attacks have complete access to the model under attack whereas blackbox attacks do not. Whitebox attacks reduced model accuracies on fake images to 0\% in all cases except ResNet with FGSM. Still, blackbox attacks resulted in less than 27\% accuracy on perturbed fake images. Furthermore, CW-$\textit{L}_{2}$ was more effective than FGSM in all cases except the blackbox attack on VGG. We suspect CW-$\textit{L}_{2}$ overfits to the ResNet model in this case.

\section{Regularization as a Defense}
\subsection{Lipschitz Regularization}
Lipschitz regularization, introduced in \cite{regularization}, constrains the gradient of the detector with respect to the input data. We achieve this by training the model using an augmented loss function involving the $\textit{L}_{2}$ norms of the logit gradients:
\begin{equation}
    J_{aug}(\textbf{x}, \textbf{y}, \theta) = J(\textbf{x}, \textbf{y}, \theta) + \frac{\lambda}{CN} \sum_{i=1}^{C} \| \nabla_x \textbf{Z}(\textbf{x})_i \|_{2}^{2}.
\end{equation}

Here, we use $J_{aug}(\textbf{x}, \textbf{y}, \theta)$ to represent the augmented loss function and $J(\textbf{x}, \textbf{y}, \theta)$ to represent the training loss function before augmentation. $\textbf{Z}(\textbf{x})_i$ represents the pre-softmax scalar output (or logit) corresponding to class $i$ for a multi-class neural network classifier. $C$ is the total number of target classes and $N$ is the dimensionality of the input vector. As before, \textbf{x} is the input vector, \textbf{y} is the corresponding true target class value and $\theta$ represents the model parameters. $\lambda$ controls the strength of the regularization term in the augmented loss function. 

Linearizing the (non-augmented) loss function provides some intuition into why this regularization can help: 
\begin{equation}
    \begin{gathered}
        J(\textbf{x}_{adv}, \textbf{y}, \theta) \approx J(\textbf{x}, \textbf{y}, \theta) + \nabla_xJ(\textbf{x}, \textbf{y}, \theta)^T(\textbf{x}_{adv} - \textbf{x}) \\
        = J(\textbf{x}, \textbf{y}, \theta) + \sum_{i=1}^{C}\frac{\partial{J}}{\partial{\textbf{Z}_{i}}}\nabla_x\textbf{Z}(\textbf{x})_{i}^T(\textbf{x}_{adv} - \textbf{x}).
    \end{gathered}
\end{equation}

As shown above, the linear approximation can be written in terms of the gradients of the detector logits with respect to the input. Then, we expect that minimizing the norm of these gradients will desensitize the loss from small perturbations, allowing the network to retain performance on inputs with adversarial perturbations. In the extreme case, if the norms of these gradients are zero, then the loss for the original unperturbed image equals the loss for the adversarially perturbed image (subject to the linear approximation).  

\subsection{Regularization Results}

Lipschitz regularization improved the detection of adversarially perturbed deepfakes by ResNet models on average. We do not report regularization results for VGG given computational constraints and the slow nature of the CW-$\textit{L}_{2}$ attack (around 2 minutes per image). We trained models with the following values for the regularization strength ($\lambda$): 5, 50, 500, and 5000. Table~\ref{attack_results_table} shows the results for all $\lambda$ values. Fig.~\ref{regularization_results} and our discussion below focus on results for the values on average and for the values that achieved the best results. Regularization did not affect the accuracy on the unperturbed test data: We observed 93.2\% accuracy for both unregularized and regularized models on average (Table~\ref{detection_results}).

In the blackbox case, unregularized models obtained an accuracy of 20.8\% for FGSM and 4.6\% for CW-$\textit{L}_{2}$ on perturbed fake images. Regularized models improved detection of perturbed images to 31.5\% for FGSM and 14.3\% for CW-$\textit{L}_{2}$ on average. In the best case, regularized models achieved an accuracy of 53.2\% for FGSM and 19.8\% for CW-$\textit{L}_{2}$ on perturbed images. 

Similarly, regularized models also performed better than unregularized models in the whitebox case. Unregularized models obtained an accuracy of 7.5\% for FGSM and 0.0\% for CW-$\textit{L}_{2}$ on perturbed fake images, as reported in section~\ref{sec_attack_results}. On average, regularized models improved detection of perturbed images to 12.5\% for FGSM and to 0.5\% for  CW-$\textit{L}_{2}$. In the best case, regularized models achieved an accuracy of 26.5\% for FGSM and 2.2\% for CW-$\textit{L}_{2}$ on perturbed images. Overall, although regularization slightly improved robustness to adversarial perturbations, the performance remains impractical for real world applications.

\section{Deep Image Prior}\label{sec_dip}
Another approach for defending against adversarial attacks is to pre-process the input to remove perturbations before feeding it to the classifier. We do this by using an unsupervised technique called Deep Image Prior (DIP) which was originally introduced in \cite{dip} for image restoration purposes such as image denoising, inpainting and super resolution. 

\begin{figure*}[t!] 
    \centering
    \includegraphics[width=0.75\textwidth]{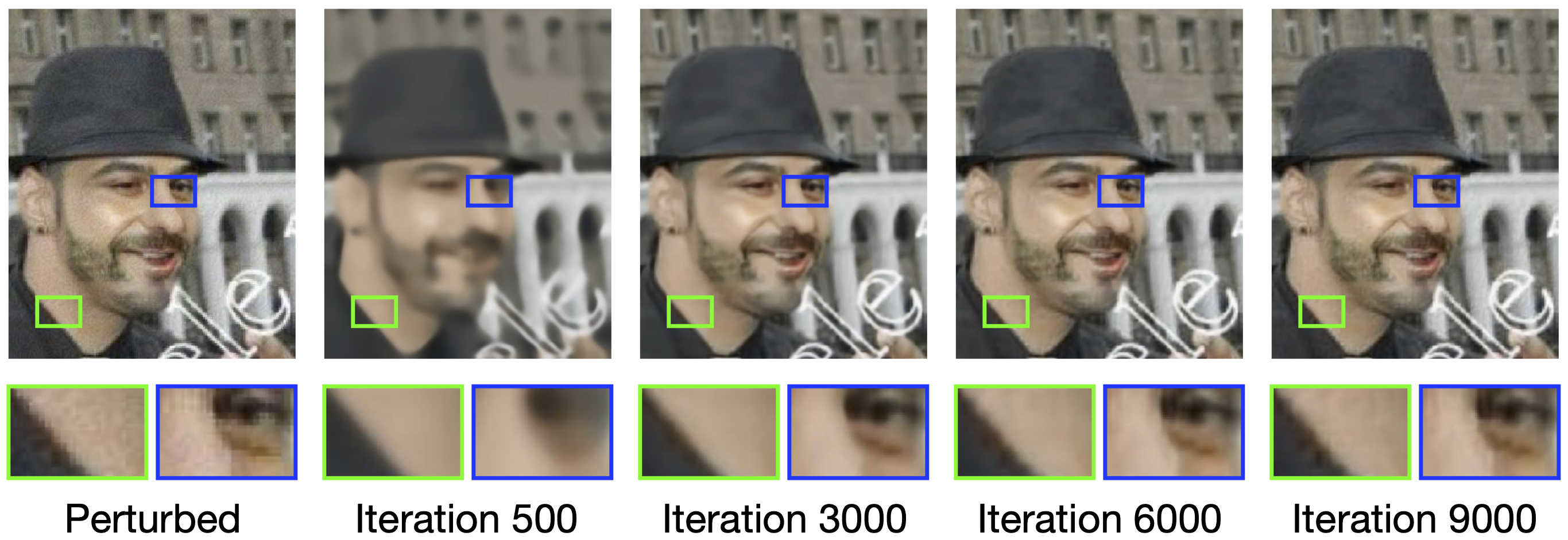}
    \caption{\textbf{Generated images during DIP optimization.} These are images generated along the optimization path for a perturbed (FGSM) fake image. The image sharpness increases along with the number of optimization iterations. However, as the images gain detail, adversarial perturbations become more apparent. The generated image at iteration 9,000 is slightly noisier than the ones at iteration 3,000 and 6,000. Image format from \cite{dip}. \textit{(Electronic zoom-in recommended).} }
    \label{dip_optimization}
\end{figure*}
\input{tables/dip_results.tex}

\input{tables/category_results.tex}

\begin{figure}[b!]
\centering
\subfloat[\textbf{Perturbed (FGSM) Fake-Wrong Image}]{\includegraphics[width=\linewidth]{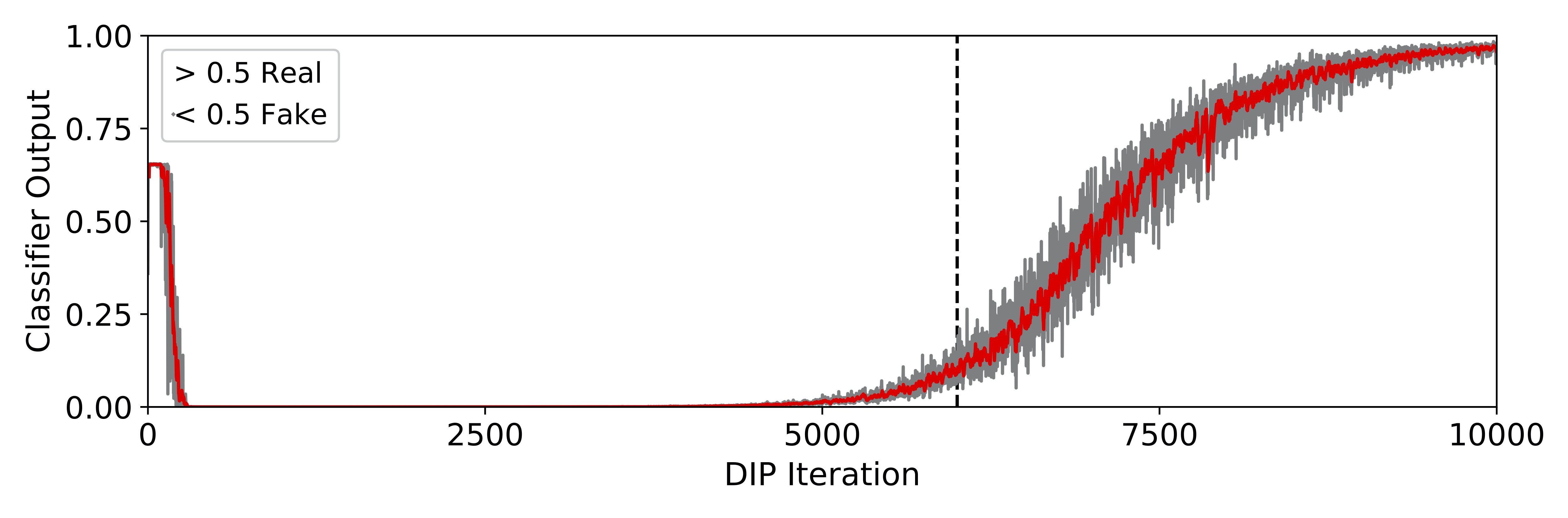}
\label{whitebox_fgsm_fake_wrong}}

\subfloat[\textbf{Perturbed (CW-$\textit{L}_{2}$) Fake-Wrong Image}]{\includegraphics[width=\linewidth]{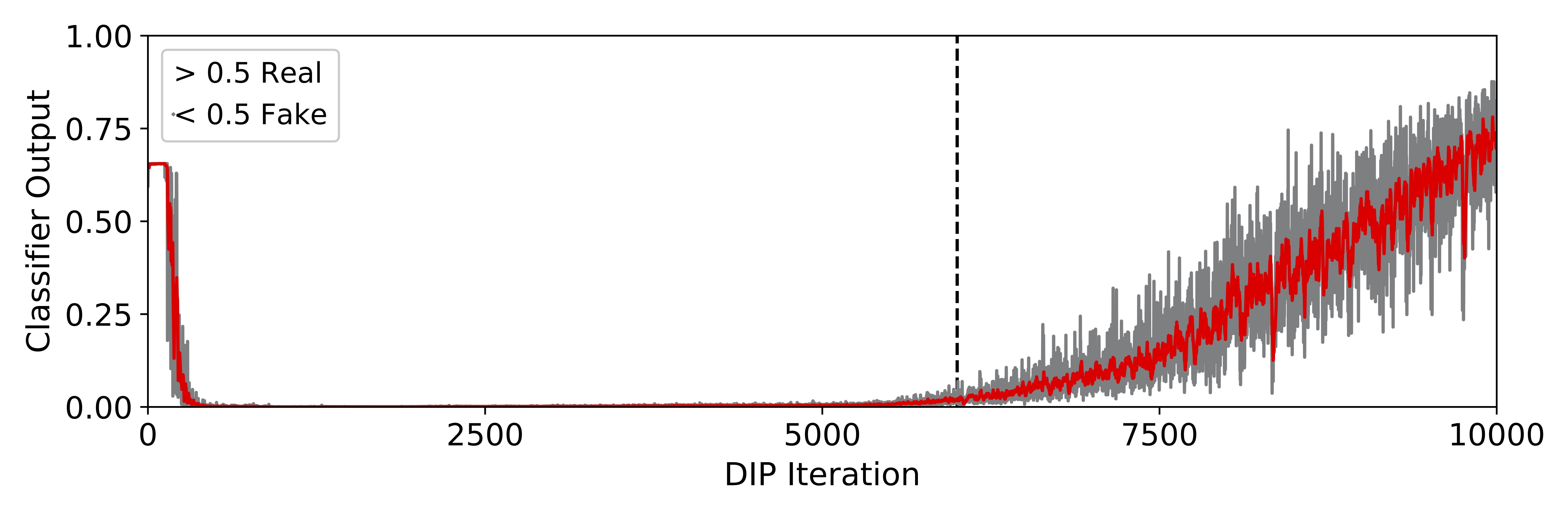}
\label{whitebox_cw_fake_wrong}}

\subfloat[\textbf{Unperturbed Fake-Correct Image}]{\includegraphics[width=\linewidth]{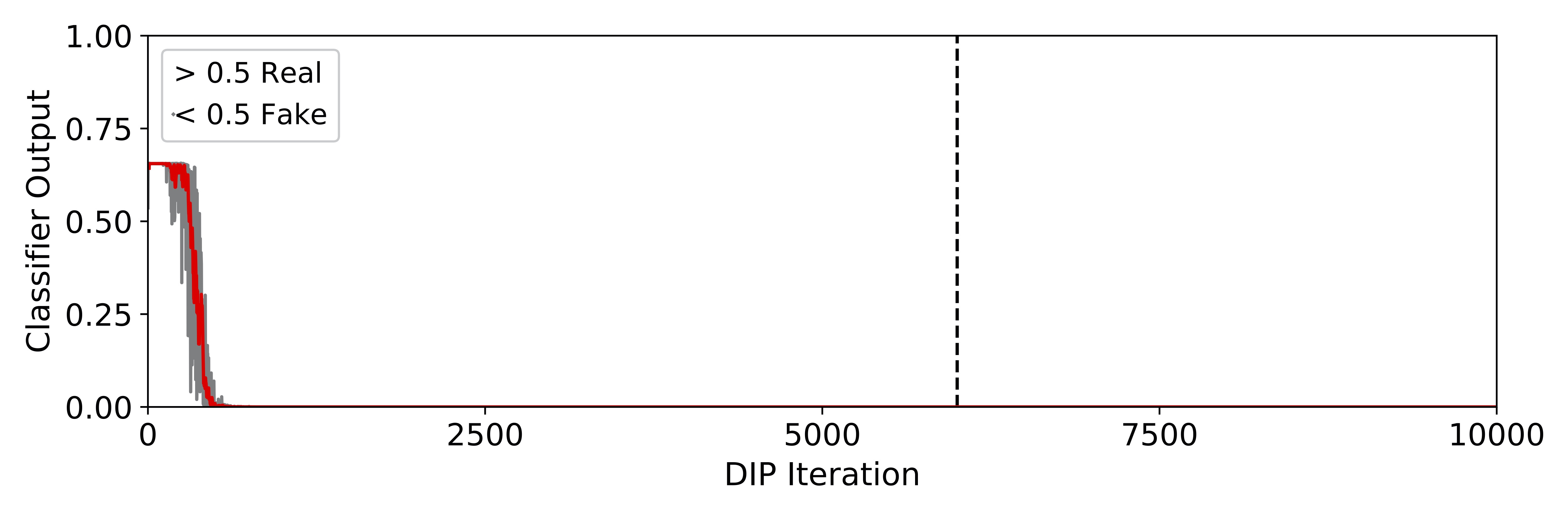}
\label{test_fake_correct}}

\subfloat[\textbf{Unperturbed Real-Correct Image}]{\includegraphics[width=\linewidth]{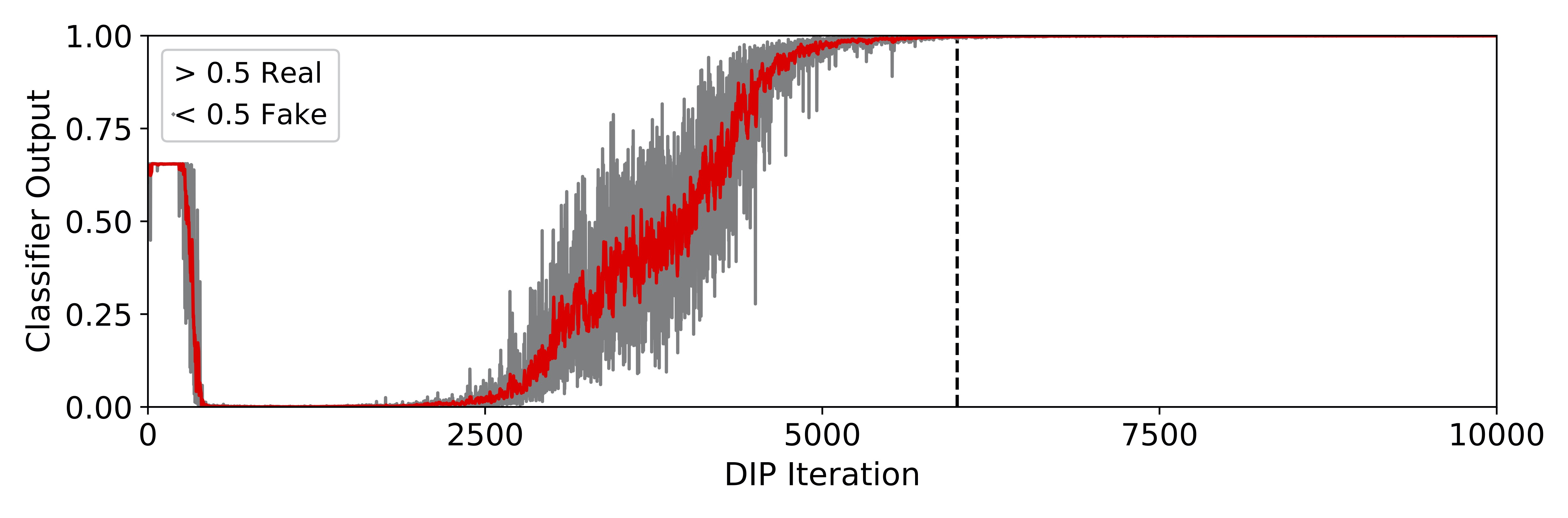}
\label{test_real_correct}}

\subfloat{\includegraphics[width=0.75\linewidth]{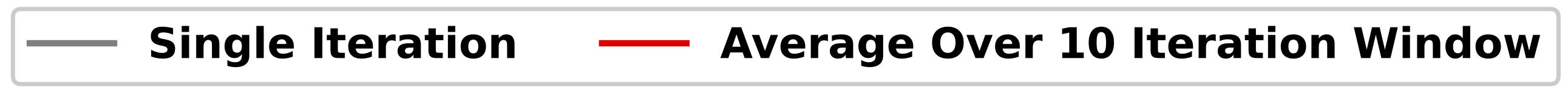}
\label{legend}}

\caption{\textbf{DIP optimization graphs.} %This figure plots the softmax output of the ResNet classifier corresponding to the real class for the generated images along the DIP optimization path for four images.
This figure plots the ResNet classifier softmax output corresponding to the real class for the generated images along the DIP optimization path for four cases. The grey curves plot the output at every iteration, while the red curves plot the average of the outputs over a window size of 10 centered at each iteration. We use 0.5 as our classification threshold for this figure. (a) and (b) correspond to fake images perturbed in a whitebox setting using FGSM and CW-$\textit{L}_{2}$ respectively. (c) and (d) correspond to unperturbed fake and real images respectively. In each case, the model classifies the generated image correctly at iteration 6,000.}
\label{dip_graphs}
\end{figure}

\subsection{Image Restoration with DIP}

This section summarizes the key ideas from the original DIP paper \cite{dip}. Let $\textbf{x}_c$ be a corrupted image (e.g. a noisy image) and $\textbf{x}$ be the ground truth uncorrupted image. Recovering $\textbf{x}$ from $\textbf{x}_c$ can be formulated as the following optimization problem: 

\begin{equation}
    \underset{x}{\min}\{ E(\textbf{x}, \textbf{x}_c) + R(\textbf{x})\}.
\end{equation}

Here, $E(\textbf{x}, \textbf{x}_c)$ represents a domain-dependent ``distance" or dissimilarity between $\textbf{x}_c$ and $\textbf{x}$. $R(\textbf{x})$ is a regularization term that represents knowledge about ground truth images. The prior knowledge from regularization is critical since recovering $\textbf{x}$ from $\textbf{x}_c$ is generally an ill-posed problem. 

We can replace $\textbf{x}$ in (9) with a surjective function $\textbf{g}: \theta \rightarrow \textbf{x}$ and optimize over $\theta$ instead:

\begin{equation}
  \underset{\theta}{\min} \{ E(\textbf{g}(\theta), \textbf{x}_c) + R(\textbf{g}(\theta))\}.
\end{equation}
The DIP technique uses a generative CNN, $\textbf{f}(\theta, \textbf{z})$, with parameters $\theta$ and random seed $\textbf{z}$ in place of $\textbf{g}(\theta)$. Through experimentation, \cite{dip} shows that the architecture of a convolutional neural network itself encodes a prior that favors natural images over corrupted ones. This allows us to get a good reconstruction even if we ignore the regularization term $R$, leading to the following optimization problem: 

\begin{equation}
    \underset{\theta}{\min} \{ E(\textbf{f}(\theta, \textbf{z}), \textbf{x}_c)\}.
\label{eq_dip_orig}
\end{equation}

In practice, if the network is optimized for too long, it learns to generate the corruptions. However, the network learns to generate ``natural" features before learning to generate the corruptions due to the prior that the CNN encodes. In other words, a good image reconstruction tends to exist somewhere along the optimization trajectory.

\subsection{Eliminating Adversarial Perturbations with DIP} 
We can use the image restoration framework described above to remove adversarial perturbations from adversarial examples. We simply replace $\textbf{x}_c$ with $\textbf{x}_{adv}$ in (\ref{eq_dip_orig}). We chose Mean Squared Error (MSE) calculated pixel-wise over the images as our dissimilarity metric, $E$. This metric was chosen since it was effective in \cite{dip} for various applications including image denoising, super resolution and JPEG compression artifact removal. Thus, we modify the DIP optimization in (\ref{eq_dip_orig}) to remove adversarial perturbations as follows: 

\begin{equation}
    \underset{\theta}{\min} \{ \MSE(\textbf{f}(\theta, \textbf{z}), \textbf{x}_{adv})\}.
\label{eq_new_dip}
\end{equation}

We propose the following deepfake defense using (\ref{eq_new_dip}). Given that an unperturbed image tends to occur somewhere along the DIP optimization trajectory, we feed the generated image at an intermediate iteration into an existing classifier. The classifier output for the generated image at the intermediate iteration is then used to make a final classification for the image. Throughout section~\ref{sec_dip}, the ``classification of the DIP defense" refers to the final classification made using this process, and ``classifier" refers to the CNN model used to obtain the classification.

We used only the ResNet model for DIP due to computational constraints as described in section~\ref{sec_dip_exp}. We also trained the classifier for an additional 10 epochs on the training dataset. For these 10 epochs, the training dataset was augmented so that approximately 40\% of it contained blurry images. This was done because the reconstructed DIP images without the perturbations tended to be slightly less sharp compared to the original images in the training and test sets. We created the blurry images by preprocessing training images using a Gaussian blur kernel with $\sigma$ values selected uniformly from the range [3.0, 5.0]. Table~\ref{detection_results} lists the performance metrics of the ResNet model trained on the augmented dataset. Table~\ref{attack_results_table} reports the adversarial attack results for this model, which are similar to the results for the model without data augmentation. 

Fig.~\ref{dip_optimization} shows a perturbed (FGSM) fake image being reconstructed using the DIP framework. This image was chosen such that the ResNet model classifies it as real. We observe that as the number of DIP optimization iterations increases, the images gain more detail. But as the image sharpness increases, the generated images also tend to include adversarial perturbations: The generated image at iteration 9,000 is slightly noiser than the ones at iterations 3,000 and 6,000. 

Fig.~\ref{whitebox_fgsm_fake_wrong} shows the classifier output for the perturbed (FGSM) fake image in Fig.~\ref{dip_optimization} along the DIP optimization path. We ignore the predictions for the first 500 iterations where the generative CNN is still learning how to produce a natural-looking image. We observe that, after this, the classifier output remains flat and close to 0 (fake) until around iteration 5,000. Following this, the classifier output increases as the generated image begins including the perturbations until it flattens out at 1 (real). Fig.~\ref{whitebox_cw_fake_wrong} shows a similar pattern for a perturbed fake (CW-$\textit{L}_{2}$) image. In contrast, for an unperturbed fake image (Fig.~\ref{test_fake_correct}), the graph flattens out at a fake prediction and never reaches a real prediction. For a real unperturbed image (Fig.~\ref{test_real_correct}), the graph reaches a real prediction much earlier than in the perturbed fake cases. In each case, the classifier predicts the correct class at iteration 6,000.

\subsection{DIP Experiments}\label{sec_dip_exp}
We performed the DIP optimization for 10,000 iterations on a total of 100 images based on the test set. Iteration 6,000 was used to obtain the classification of the DIP defense after evaluating iterations in the range of 2,500 to 7,500. We used a U-Net architecture \cite{unet} for the generative CNN, $\textbf{f}(\theta, \textbf{z})$, in the DIP optimization described in (\ref{eq_new_dip}). This architecture was used because it was shown to be effective in \cite{dip} for both denoising and removing JPEG compression artifacts from images. For the exact architecture details, we refer to our code repository linked at the end of this paper. The optimization process is slow and took approximately 30 minutes for each image using a NVIDIA Tesla K80 GPU on Google Colab. For this reason, we chose only 100 images for the experiments. 

We randomly sampled 10 images each from the following 2 categories for both perturbed FGSM and CW-$\textit{L}_{2}$ images in blackbox and whitebox settings (80 images total):
\begin{itemize}[leftmargin=*]
    \item \textit{Perturbed Fake-Wrong}: A perturbed fake image that the classifier predicts as real. 
    \item \textit{Perturbed Fake-Correct}: A perturbed fake image that the classifier predicts as fake. 
\end{itemize}
In addition, we sampled 10 images each from the following 2 categories for unperturbed images (20 images total): 
\begin{itemize}[leftmargin=*]
    \item \textit{Unperturbed Fake-Correct}: An unperturbed fake image that the classifier predicts as fake. 
    \item \textit{Unperturbed Real-Correct}: An unperturbed real image that the classifier predicts as real. 
\end{itemize}
All images were sampled such that the ResNet model (trained on the augmented dataset) obtained a correct prediction on the unperturbed versions of the images.

\subsection{DIP Results}

%The DIP defense successfully classified both perturbed and unperturbed fake images as well as unperturbed real images. Table~\ref{dip_results} reports the overall performance of the DIP defense across all 100 images. Table~\ref{category_results} includes the accuracy of the DIP defense for each category of images. We report results using classification thresholds of 0.5 and 0.25; this threshold was used to obtain a prediction from the classifier output on the DIP generated image at iteration 6,000. 

%The DIP defense achieved 97.0\% accuracy and 99.2\% AUROC for both classification thresholds. Varying the threshold reveals the tradeoff between incorrectly predicting more real images as fake (false positives) and more fake images as real (false negatives). For deepfake detection, false positives may be less of a problem than false negatives.

%Using a threshold of 0.5 yielded 100\% recall for both unperturbed and perturbed fake images. However, this threshold resulted in only 70\% recall for real images. On the other hand, using a threshold of 0.25 improved the recall for real images to 90\%. But this also caused false negatives to increase; specifically, the accuracy in the whitebox FGSM perturbed fake-wrong category decreased from 100\% to 80\%. 

We report DIP results  using classification thresholds of 0.5 and 0.25. Table~\ref{dip_results} reports the overall performance of the DIP defense across all 100 images while Table~\ref{category_results} includes the accuracy of the DIP defense for each category of images. The tables also include a baseline performance from the classifier without the DIP defense.

The DIP defense achieved 95\% on perturbed deepfakes that fooled the original detector (Perturbed Fake-Wrong), while retaining 98\% accuracy in other cases (Real-Correct and Fake-Correct) with a 0.25 threshold. Overall, on the 100 image subsample, the defense obtained 97.0\% accuracy and 99.2\% AUROC for both classification thresholds. Varying the threshold reveals the tradeoff between incorrectly predicting real images as fake (false positives) and fake images as real (false negatives). For deepfake detection, false positives are generally less of a problem than false negatives.

Using a threshold of 0.5 yielded 100\% recall for both unperturbed and perturbed fake images. But this threshold resulted in only 70\% recall for real images. On the other hand, using a threshold of 0.25 improved the recall for real images to 90\%, but also reduced recall for fake images to 97.8\%. Specifically, the accuracy in the whitebox perturbed (FGSM) fake-wrong category decreased from 100\% to 80\%. 

%It also retained 100\% accuracy in the Perturbed Fake-Correct categories, while improving performance in the Fake-Wrong categories. 
%Table~\ref{category_results} reports the accuracy of the classifier after using DIP procedure for each category of images using classification thresholds of 0.5 and 0.25. Table~\ref{dip_results} lists additional performance metrics over all 100 images. We see in Table Y that using a classification threshold of 0.5, the classifier was able to recall 100\% of fake images whether perturbed or not. However, it only recalled 70\% of the real images. For a deepfake detector, predicting real images as fake (false positives) is less of a problem compared to predicting fake images as real (false negatives). The high model AUROC (99.2\%) suggests that lowering the threshold might help improve real recall and thus reduce the number of false positives. With a classification threshold of 0.25, the model improved the real recall to 90\%. Of course, this also caused the false negatives to increase: Lowering the threshold from 0.5 to 0.25 caused the accuracy for the whitebox FGSM perturbed fake-wrong category to go down from 100\% to 80\% (Table~\ref{category_results}) and also brought down the overall fake recall down from 100\% to 97.8\% (Table~\ref{dip_results}).

\section{Discussion and Limitations}
Our results demonstrate that adversarial perturbations can enhance deepfakes, making them significantly more difficult to detect. Lipschitz regularization made the CNNs more robust to adversarial perturbations in general. However, the performance boost from regularization alone may not be enough for practical use in deepfake detection. This was especially true in the whitebox CW-$\textit{L}_{2}$ setting where even the regularized model only classified 2.2\% of the perturbed fake images correctly. The DIP defense shows more promising results. It achieved a recall of 97.8\% 
for perturbed and unperturbed fake images using a classification threshold of 0.25 (Table~\ref{dip_results}). Furthermore, the DIP defense retained at least 90.0\% of the classifier's performance on real images using the same threshold value. 

%The DIP defense improves robustness against adversarially perturbed deepfakes for the 100 images tested. However, we emphasize that additional experiments are needed, especially to explore using DIP as a defense for adversarial attacks in other domains. For example, deepfake classifiers only need to be robust to adversarial perturbations for one class of images (the fake class), while in other domains robustness to adversarial attacks on more than one classes is likely important. Another limitation of the DIP defense is the long time it takes to process a single image; as discussed in the DIP section, 10,000 iterations took about 30 minutes on a NVIDIA Tesla K80 GPU. We can reduce this computation time by fixing an iteration to obtain the final DIP classification ahead of time. For example, we observed that iteration 6,000 worked well in this work; thus it would have sufficed to only optimize for 6,000 iterations. Nevertheless, computation time may limit the practicality of the DIP defense in situations where there are resource constraints or where many images need to be processed in real time. Further work is needed to find more efficient methods to improve robustness of deepfake detectors against adversarial attacks. Overall, through this work we hope to inspire additional research to improve robustness of deepfake detectors to adversarial perturbations. 

While the DIP defense showed success for deepfake detection on the 100 images tested, we emphasize that additional experiments would be required to demonstrate success on adversarial attacks in other domains. For example, deepfake classifiers only need to be robust to adversarial perturbations for one class of images (the fake class), while in other domains, robustness to adversarial attacks on more than one class may be important. Another limitation of the DIP defense is the time it takes to process a single image. As described in section~\ref{sec_dip_exp}, each image took a little under 30 minutes to process on a NVIDIA Tesla K80 GPU. This may limit the practicality of the defense in situations where there are resource constraints or where many images need to be processed in real time. Future work involves finding more efficient methods for improving deepfake detector robustness to adversarial perturbations.

%Future work involves finding more efficient methods to remove perturbations. Overall, through this work, we hope to inspire additional research to improve robustness of deepfake detectors to adversarial perturbations.  %Still, we can bring down computation time for DIP by fixing an iteration to obtain the final DIP classification ahead of time. For example, in this work since we observe that iteration 6,000 worked on the dataset of 100 images, when processing future images, we need only optimize for 6,000 iterations.

\section*{Acknowledgment}
The authors sincerely thank Professor Bart Kosko for his feedback and guidance throughout this work. We also acknowledge the USC Center for AI in Society's Student Branch for providing us with computing resources. 

\section*{Code Availability}
Code and additional architecture details are available at: \href{https://github.com/ApGa/adversarial\_deepfakes}{\textit{https://github.com/ApGa/adversarial\_deepfakes}}.

%The following repository provides implementations for our experiments and additional model architecture detail: \href{https://github.com/ApGa/adversarial\_deepfakes}{\textit{https://github.com/ApGa/adversarial\_deepfakes}}.

\bibliographystyle{IEEEtran}
\bibliography{IEEEabrv, references}

\end{document}

%% file: tables/detection_results.tex
\begin{table*}[b!]
\caption{Unperturbed Data Results}
\centering
\begin{tabular}{p{4cm}p{1cm}p{1cm}p{0.2cm}p{1cm}p{1cm}p{0.2cm}p{1cm}p{1cm}}
\toprule
\multirow{2}{*}{Model} & \multirow{2}{*}{Accuracy} & \multirow{2}{*}{AUROC} & & \multicolumn{2}{c}{Fake} & & \multicolumn{2}{c}{Real} \\ 
& & & & Precision & Recall & & Precision & Recall \\ 
\toprule
VGG & 99.7\% & 99.9\% && 99.8\% & 99.7\% && 99.7\% & 99.8\% \\
ResNet & 93.2\% & 97.9\% && 91.5\% & 95.4\% && 95.2\% & 91.1\% \\
ResNet (Regularized: $\lambda$=5) & 95.0\% & 99.5\% && 91.5\% & 99.3\% && 99.2\% & 90.8\% \\
ResNet (Regularized $\lambda$=50) & 94.1\% & 98.6\% && 96.6\% & 91.4\% && 91.8\% & 96.8\% \\
ResNet (Regularized $\lambda$=500) & 87.5\% & 92.3\% && 85.3\% & 90.6\% && 89.9\% & 84.4\% \\
ResNet (Regularized $\lambda$=5000) & 96.2\% & 99.1\% && 98.0\% & 94.2\% && 94.5\% & 98.1\% \\
ResNet (Average Regularized) & 93.2\% & 97.4\% && 92.9\% & 93.9\% && 93.9\% & 92.5\% \\
ResNet (Data Augmented) & 98.7\% & 99.9\% && 98.5\% & 98.9\% && 98.9\% & 98.5\% \\
\bottomrule
\multicolumn{9}{l}{\textit{Note: Unperturbed results listed for a test dataset containing 1,250 images each of real and fake classes.}} \\
\end{tabular}
\label{detection_results}
\end{table*}

%% file: tables/attack_results.tex
\begin{table*}[b!]
\caption{Adversarial Attack Results}
\centering
\begin{tabular}{p{4cm}p{1.9cm}p{0.2cm}p{1cm}p{1cm}p{0.2cm}p{1cm}p{1cm}}
\toprule
\multirow{2}{*}{Model} & \multirow{2}{*}{Unperturbed}
&& \multicolumn{2}{c}{Perturbed: FGSM} && \multicolumn{2}{c}{Perturbed: CW-$\textit{L}_{2}$} \\ 
& & & Blackbox & Whitebox && Blackbox & Whitebox \\
\toprule
VGG & 99.7\% && 8.9\% & 
0.0\% && 26.6\% & 0.0\% \\
ResNet & 95.4\% && 20.8\% & 7.5\% && 4.6\% & 0.0\% \\
ResNet (Regularized $\lambda$=5) & 99.3\%$^*$ && 42.2\% & 26.5\%$^*$ && 14.5\% & 0.0\% \\
ResNet (Regularized $\lambda$=50) & 91.4\% && 17.8\% & 12.7\% && 6.2\% & 0.0\% \\
ResNet (Regularized $\lambda$=500) & 90.6\% && 53.2\%$^*$ & 9.0\% && 19.8\%$^*$ & 0.0\% \\
ResNet (Regularized $\lambda$=5000) & 94.2\% && 12.8\% & 1.9\% && 16.7\% & 2.2\%$^*$ \\
ResNet (Average Regularized) & 93.9\% && 31.5\% & 12.5\% && 14.3\% & 0.5\% \\
ResNet (Data Augmented) & 98.9\% && 17.0\% & 2.2\% && 3.8\% & 0.1\% \\
\bottomrule
\multicolumn{8}{l}{\textit{Note: Adversarial attacks conducted using only the fake images. $^*$Best regularized performances.}} \\ 
\end{tabular}
\label{attack_results_table}
\end{table*}

%% file: tables/dip_results.tex
\begin{table*}[t!]
\caption{DIP Defense Results Overall}
\centering
\begin{tabular}{p{1.5cm}p{1cm}p{1cm}p{0.2cm}p{1cm}p{1cm}p{0.2cm}p{1cm}p{1cm}}
\toprule
\multirow{2}{*}{\shortstack[l]{Classifier\\Threshold}} & \multirow{2}{*}{Accuracy} & \multirow{2}{*}{AUROC} & & \multicolumn{2}{c}{Fake} & & \multicolumn{2}{c}{Real} \\ 
& & & & Precision & Recall & & Precision & Recall \\ 
\toprule
0.50 & 97.0\% & 99.2\% && 96.8\% & 100\% && 100\% & 70.0\% \\
\midrule
0.25 & 97.0\% & 99.2\% && 98.9\% & 97.8\% && 81.8\% & 90.0\% \\
\midrule
\midrule
Baseline & 60.0\% & 41.9\% && 100.0\% & 55.5\% && 20.0\% & 100.0\% \\
\bottomrule
\multicolumn{9}{l}{\textit{Note: DIP Results based on 100 images subsampled according to section V-C.}} \\ 
\end{tabular}
\label{dip_results}
\end{table*}

%% file: tables/category_results.tex
\begin{table*}[t!]
\caption{DIP Defense Results by Category}
\centering
\begin{tabular}{p{1.5cm}p{1.5cm}p{1.5cm}p{1.5cm}p{0.2cm}p{1.5cm}p{1.5cm}p{0.2cm}p{1.5cm}p{1.5cm}}
\toprule
\multirow{2}{*}{\shortstack[l]{Classifier\\Threshold}} & \multirow{2}{*}{Attack} &  
\multicolumn{2}{c}{Unperturbed}
& & \multicolumn{2}{c}{Blackbox: Perturbed} & & \multicolumn{2}{c}{Whitebox: Perturbed} \\ 
& & Fake-Correct & Real-Correct & & Fake-Wrong & Fake-Correct & & Fake-Wrong & Fake-Correct \\ 
\toprule
\multirow{2}{*}{0.50} & FGSM & \multirow{2}{*}{100\%} & \multirow{2}{*}{70\%} && 100\% & 100\% && 100\% & 100\% \\
& CW-$\textit{L}_{2}$ & & & & 100\% & 100\% && 100\% & 100\% \\
\midrule
\multirow{2}{*}{0.25} & FGSM & \multirow{2}{*}{100\%} & \multirow{2}{*}{90\%} && 100\% & 100\% && 80\% & 100\% \\
& CW-$\textit{L}_{2}$ & & & & 100\% & 100\% && 100\% & 100\% \\
\midrule
\midrule
\multirow{2}{*}{Baseline} & FGSM & \multirow{2}{*}{100\%} & \multirow{2}{*}{100\%} && 0\% & 100\% && 0\% & 100\% \\
& CW-$\textit{L}_{2}$ & & & & 0\% & 100\% && 0\% & 100\% \\
\bottomrule
\multicolumn{10}{l}{\textit{Note: DIP Results based on 100 images subsampled according to section V-C.}} \\ 
\end{tabular}
\label{category_results}
\end{table*}